\documentclass{cup-ino}
\usepackage{xparse}         
\makeatletter
\let\origHat\^              
\RenewDocumentCommand{\^}{o}
  {%
    \IfNoValueTF{#1}        
      {\textasciicircum}    %
      {\origHat{#1}}        
  }
\makeatother
\usepackage[utf8]{inputenc}
\usepackage[T1]{fontenc}
\usepackage{lmodern}
\usepackage{textcomp}
\usepackage{csquotes}
\usepackage{appendix}
\usepackage{authblk}
\usepackage{longtable}
\usepackage{blindtext,alltt}
\usepackage{float}
\usepackage{multirow}
\usepackage[backend=biber]{biblatex}

\newenvironment{keywords}{\begin{center}\bfseries Keywords:}{\end{center}}
\title{Loop2Net: Data-Driven Generation and Optimization of Airfoil CFD Meshes from Sparse Boundary Coordinates}

\author{Lushun Fan, Yuqin Xia, Jun Li, Karl Jenkins}

\begin{document}
\maketitle

\begin{abstract}
In this study, an innovative intelligent optimization system for mesh quality is proposed, which is based on a deep convolutional neural network architecture, to achieve mesh generation and optimization. The core of the study is the Loop2Net generator and loss function, it predicts the mesh based on the given wing coordinates. And the model's performance is continuously optimised by two key loss functions during the training. Then discipline by adding penalties, the goal of mesh generation was finally reached.
\end{abstract}

\begin{keywords}
Data-Driven
\end{keywords}


\section{Introduction}

In computational fluid dynamics (CFD) for airfoil design, generating high-quality meshes is essential for improving simulation accuracy and engineering productivity. Traditional meshing workflows often rely on commercial software and expert manual adjustments, which become especially cumbersome when only sparse coordinate information is available, such as the geometric outline of the airfoil. In such cases, the transformation from boundary points to simulation-ready meshes typically requires significant preprocessing and manual effort. \\

To address this challenge, this study introduces a data-driven framework that enables the automatic generation of dense CFD mesh coordinates from a limited set of ordered boundary inputs. The proposed method utilizes a neural network model trained to infer the spatial distribution of fluid mesh nodes directly, offering a streamlined alternative to conventional meshing pipelines. \\

Several core difficulties are associated with this task. These include the sparsity of the input, the need to constrain output nodes within the valid fluid region surrounding the airfoil, and the requirement that the resulting mesh match professional standards of distribution and structural quality. To meet these demands, our framework integrates targeted data preprocessing, a carefully designed network architecture, composite loss functions, and a progressive training strategy. \\

Experimental results demonstrate that the model can successfully generate muliti hundred nodes surrounding an airfoil from only 35 input coordinates, achieving strict adherence to the defined boundary. Visual and quantitative evaluations show that the predicted node distributions align well with those produced by commercial software in terms of coverage, density, and spatial regularity. The generated meshes exhibit zero intrusion into the airfoil interior, reduced Chamfer distance from the reference grid, and a consistent node dispersion pattern. This method significantly lowers the entry barrier for high-quality two-dimensional aerodynamic mesh generation and provides a highly efficient toolchain suitable for simulation, design iteration, and parametric optimization.

\section{Related Work}
\subsection{Mesh Optimization}
Mesh generation technology is an important part of the computational fluid dynamics research system, and the report released by NASA \cite{Slotnick2014CFDAerosciences} in 2014 clearly states that geometric modeling and mesh generation, as one of the six key technology areas, has become an important direction for future CFD research, which also makes mesh optimization methods using deep learning methods a research hotspot. This has also prompted the mesh optimization method based on deep learning to become a research hotspot in the current academic community. It is worth pointing out that deep learning has shown broad application prospects in the field of fluid dynamics: in partial differential equation solving, the deep Galyokin method proposed by Sirignano et al\cite{Sirignano2018DGM:Equations}, and the physically constrained neural network developed by Thuerey's team\cite{Thuerey2018DeepFlows} provide a new paradigm for the calculation of the flow field; in the field of grid optimization, the progressive grid optimization theory proposed by \cite{Hoppe1993MeshOptimization}, and the progressive grid optimization theory proposed by Lin et al\cite{Lin2022AMeshes}, are the most popular grid optimization methods in the world. , the adaptive mesh generation technique based on deep reinforcement learning developed by Lin et al\cite{Lin2013NetworkNetwork}, and the geometrically conformal optimization algorithm proposed by Knupp\cite{Knupp2022GeometricParadigm} are a series of achievements that continue to drive the technological innovation in this field.\\

\subsection{NNs for mesh generation and optimization}
In the field of Computational Fluid Dynamics (CFD) and engineering optimisation, the generation of high-quality surface/body meshes is always the most time-consuming and experience-dependent process before numerical simulation. Traditional algorithms such as Delaunay triangulation, Advancing-Front, O-grid, etc\cite{Chen2022AEvaluation}, although robust, are difficult to fully automate for a large number of variant geometries and still require a lot of parameter tuning. In recent years, deep learning breakthroughs in geometry processing have provided new ideas for ‘learning mesh generation’: outputting point clouds or meshes directly from geometric profiles or implicit representations via end-to-end networks not only reduces manual intervention, but also significantly accelerates pre-processing during multiple iterations of CFD calls. With deep learning breakthroughs on non-Euclidean data, researchers have begun to try to get the network to ‘directly learn’ to infer meshes from geometric or physical boundaries, which opens up the possibility of completely freeing up the pre-processing chain.\\

Deep learning breakthrough in point cloud reconstruction without topological constraints. FoldingNet\cite{Yang2018FoldingNet:Deformation} maps regular 2D templates to target surfaces using a ‘2D mesh origami’ decoder, enabling end-to-end point cloud self-encoding. AtlasNet\cite{Groueix2018AtlasNet:Generation} further extends this idea to the stitching of multiple learnable 2D patches, generating point clouds or triangles of arbitrary resolution, with good adaptability to complex topologies. However, both of them do not explicitly constrain the mesh flow structure, which makes it difficult to directly meet the strict requirements of CFD on cell orthogonality and boundary layer thickness.\\

After that, to compensate for the lack of topology, the researchers proposed a mesh convolution-meshable path. MeshCNN\cite{2019MeshCNN:Edge} defines convolution and pooling operations on triangular mesh edges, and is able to learn both geometric and topological features. Subsequently, Neural Mesh Flow directly generates diabatic meshes with the help of a differentiable diffeomorphic flow, which improves the consistency and physical availability of the generated results\cite{Gupta2020NeuralFlows}. These efforts have laid the technical groundwork for derivable mesh generation, but are still far from the local anisotropy encryption and boundary layer detail required for CFD. \\

At the same time, reinforcement learning and graph neural networks are introduced for adaptive mesh optimisation. MeshDQN models mesh coarsening as a Markov decision process, and iterative deletion of nodes with a dual DQN can drastically reduce the total number of cells while maintaining the target aerodynamic coefficients\cite{Lorsung2022MeshDQN:Dynamics}. In more general finite element adaptive scenarios, Yang et al.\cite{Yang2023ReinforcementRefinement} and Li et al.\cite{Freymuth2023SwarmRefinement} consider AMR as a local sequential decision-making problem, and achieve better mesh allocation strategies than the classical heuristics through single-intelligent and population reinforcement learning, respectively, and verify the generalisation ability on multi-physics benchmarks.\\ 

In addition to purely data-driven approaches, physical information has also been used to guide mesh generation.Physics-Informed Neural Networks (PINNs) proposed by Raissi et al.\cite{Raissi2019Physics-informedEquations} write PDE residuals directly into the loss function, providing a general framework for embedding physical conservation in deep networks. Based on this, Chen et al.\cite{Chen2022AnNetworks} proposed an end-to-end generation method for 2D structured meshes by considering PINN as a global optimiser and imposing boundary conditions and smoothing constraints on the mapping function, which significantly reduces the mesh distortion while maintaining the consistency of the flow pattern. Despite the encouraging results, the method is still difficult to control carefully the wall orthogonality and its applicability in high Reynolds number flow fields has not been verified.

\subsection{Conclusion}
Taken together, current research has seen a variety of parallel but not converged routes, such as point cloud methods for generation focusing on the overall geometry but lacking constraints, PINN which writes physical conditions into the loss but has no control over the boundary conditions, and reinforcement learning which emphasises local encryption but relies on the initial mesh.
This research fulfils the gap in the field of generating a satisfying mesh layout directly from the airfoil contours.

\section{Methodology}

\subsection{Overview}
\begin{figure}[!h]
    \centering
    \includegraphics[width=0.9\linewidth]{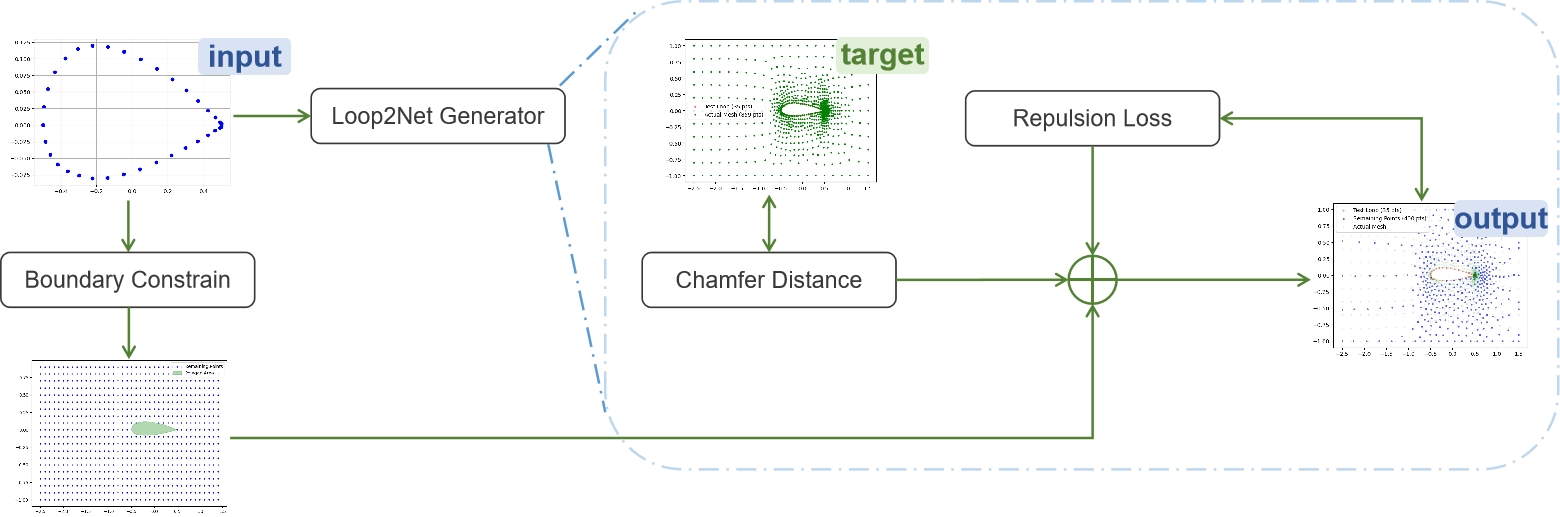}
    \caption{Mesh Generation and Optimization. The process begins with the input of airfoil coordinates, which are mapped to a closed polygon to define the wing area. The Loop2Net generator predicts a dense mesh grid based on the input coordinates. The model is trained using self-loss, which ensures internal consistency of the predicted mesh, and alignment-loss, which compares the predicted mesh to the true grid and optimizes its alignment and point distribution.}
    \label{fig:cnn_flow}
\end{figure}

This project focuses on generating mesh grids representing the fluid distribution around an airfoil, shown as Figure \ref{fig:cnn_flow}. The process begins with inputting the airfoil coordinates, which are used to predict a dense set of mesh points that represent the surrounding fluid. Initially, the airfoil coordinates are mapped to a closed polygon, ensuring that all generated mesh points remain within the defined boundary, preventing any out-of-bounds generation. \\

The core of the model involves the Loop2Net generator, which predicts the mesh grid based on the given airfoil coordinates. During training, the performance of model is continuously optimized through two key loss functions. Alignment Loss, like Chamfer Distance, which minimizes the discrepancy between the predicted and actual mesh grids. The other one is Self Loss, such as Repulsion Loss, which measures the loss value within self, allowing the model to refine its internal consistency. \\

\subsection{Dataset}
The NACA series airfoils were used as the baseline dataset at the data processing, and each sample contains two associated files: a .dat file to record the geometric coordinates of the airfoils and an .msh file to store the reference mesh. The preprocessing process normalises the geometry through a function that translates the chord to the leading edge zero coordinate system and then extracts the node coordinates from the grid file. This is followed by operations such as resampling the reference mesh with a fixed number of points to fit the fully connected network output dimensions, and implementing an optional zero-mean unit-variance normalisation to remove the effect of airfoil size differences on the model.\\

\subsection{Loop2Net}
\subsubsection{Generator}
\begin{figure}
    \centering
    \includegraphics[width=0.5\linewidth]{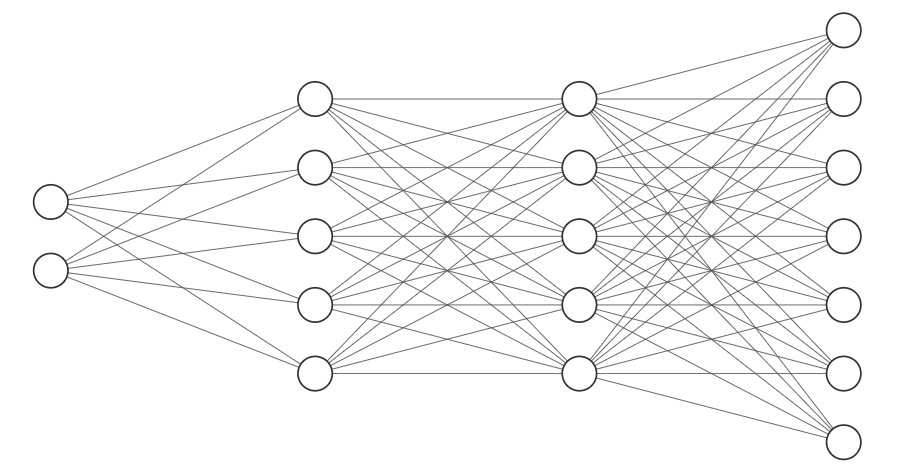}
    \caption{Fully connected layers}
    \label{fig:linear_layer}
\end{figure}

The Loop2MeshNet consists of three fully connected layers, shown as Figure \ref{fig:linear_layer}. The first two layers transform the flattened input into increasingly abstract representations, with each layer followed by a ReLU activation function. The output from the second hidden layer is then passed through a third layer to generate an output vector corresponding to the target number of generated mesh points. The final output is reshaped into a tensor of shape \textbf{B x N x 2}, where N is the target number of mesh points and each point has two coordinates (x, y). \\

The network uses Xavier \cite{Glorot2021UnderstandingNetworks} uniform initialization for its weights to ensure stable learning throughout the training process. The forward pass processes the airfoil boundary coordinates and generates the mesh grid prediction, which is refined during training to closely match the ground truth mesh grid.

\subsubsection{Loss Function and Training}
\textbf{Chamfer Distance Loss} is a geometric alignment loss to measure the similarity between the predicted mesh points and the ground-truth mesh distribution. Let $\mathcal{P} = \{p_i\}_{i=1}^{N_p}$ be the set of $N_p$ predicted points and $\mathcal{G} = \{g_j\}_{j=1}^{N_g}$ be the set of $N_g$ ground truth mesh. The Chamfer Distance is defined as the Equation \ref{eq:chamfer}

\begin{equation}
\mathcal{L}_{\text{Chamfer}}(\mathcal{P}, \mathcal{G}) = \sum_{p \in \mathcal{P}} \min_{g \in \mathcal{G}} \|p - g\|_2^2 + \sum_{g \in \mathcal{G}} \min_{p \in \mathcal{P}} \|g - p\|_2^2
\label{eq:chamfer}
\end{equation}

This loss penalizes both missing coverage, ensuring each ground-truth point is matched by a nearby prediction, and over-prediction, encouraging predicted points to align with the ground truth. We implement this loss in a fully batched manner by computing all pairwise squared distances and performing per-point minimum pooling, followed by averaging over the batch. \\


\textbf{Repulsion Loss} promotes spatial dispersion among the predicted nodes to discourage excessive clustering of predicted mesh points. It is defined as the inverse of the average pairwise L2 distance among all predicted nodes as the Equation \ref{eq:repulsion}, where $\|p_i - p_j\|_2$ denotes the L2 distance between node \(i\) and node \(j\), and the summation includes self-pairs (\(i = j\)), which contribute zero but are negligible due to averaging. 

\begin{equation}
\mathcal{L}_{\text{Repulsion}} = \left( \frac{1}{N_p^2} \sum_{i=1}^{N_p} \sum_{j=1}^{N_p} \|p_i - p_j\|_2 \right)^{-1}
\label{eq:repulsion}
\end{equation}

Intuitively, when the predicted nodes are tightly clustered, the average pairwise distance becomes small, causing the repulsion loss to increase and thus penalize such configurations. Moreover, a small constant is added inside the square root to ensure numerical stability.

\subsection{Evaluator}

To evaluate the distributional similarity between the predicted mesh and the ground truth CFD mesh, we adopt a density-based comparison framework based on kernel density estimation (KDE) and Kullback--Leibler (KL) divergence. KDE is employed to estimate continuous probability density functions from the predicted and reference point clouds by applying Gaussian smoothing over a uniformly sampled spatial grid. Let $P(x, y)$ and $Q(x, y)$ denote the estimated probability densities of the predicted and reference (ground truth) meshes, respectively, defined over a fixed 2D window $[x_{\min}, x_{\max}] \times [y_{\min}, y_{\max}]$. The KL divergence between the two distributions is computed as the Equation \ref{eq:DKL}. where $\epsilon$ is a small constant added to prevent division by zero and ensure numerical stability. The KL divergence measures how much the predicted distribution $P$ diverges from the reference distribution $Q$. \\

\begin{equation}
D_{\mathrm{KL}}(P \,\|\, Q) = \sum_{x,y} P(x, y) \log\left(\frac{P(x, y)}{Q(x, y) + \epsilon}\right)
\label{eq:DKL}
\end{equation}

A lower $D_{\mathrm{KL}}$ value indicates a higher similarity between the predicted and reference point distributions, implying better alignment in terms of spatial extent and local density. Conversely, a higher KL value suggests that the predicted mesh deviates from the target distribution, potentially due to overconcentration, sparsity, or spatial displacement. This metric offers an interpretable and quantitative means to evaluate the fidelity of generated mesh point distributions, particularly suitable for tasks involving structured spatial layouts.

\section{Experiment}

\subsection{Dataset}
\textbf{Description:} The airfoil meshes are generated based on the NACA series, and the original fine mesh and the coarse mesh are generated manually by ICEM and correspond to each other. Unstructured meshes were used for all meshes and the quality of the mesh was deliberately controlled to be fine and coarse for comparison in subsequent tests. \\

\textbf{Preprocessing:} involved mapping the coordinate points to a closed polygon that defines the wing area, ensuring that all generated points remain within the specified boundaries. Subsequently, the real grid data was organized into pairwise inputs with the loop coordinates. These pairwise inputs are then used for model training, facilitating the correction of the output through the comparison with the true grid data. Additionally, due to the requirement that stack expects each tensor to have consistent size during neural network training and testing, all target mesh data were uniformly upsampled to a fixed number of nodes. In this study, the upsampled resolution was set to 1500 points.

\subsection{Result}

\subsubsection{Standardisation}

\begin{figure}[]
    \centering
    \subfloat[]{
        \includegraphics[width=0.5\linewidth]{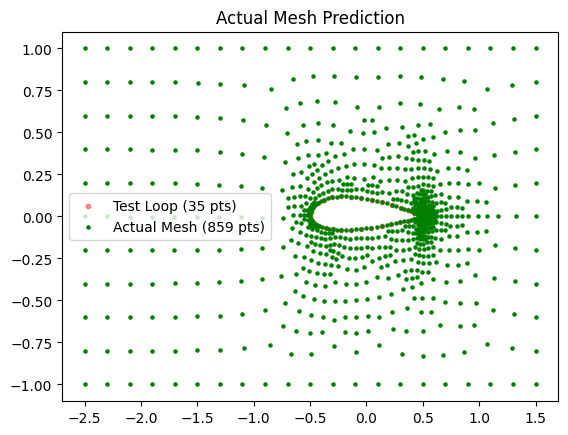}
        \label{fig:2220_ori_target}
    }
    \subfloat[]{
        \includegraphics[width=0.5\linewidth]{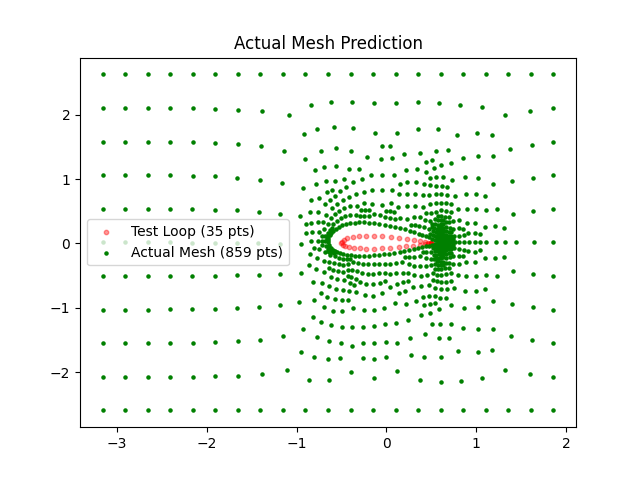}
        \label{fig:2220_stand_target}
    }
    \vspace{-2pt}
    \subfloat[]{
        \includegraphics[width=0.5\linewidth]{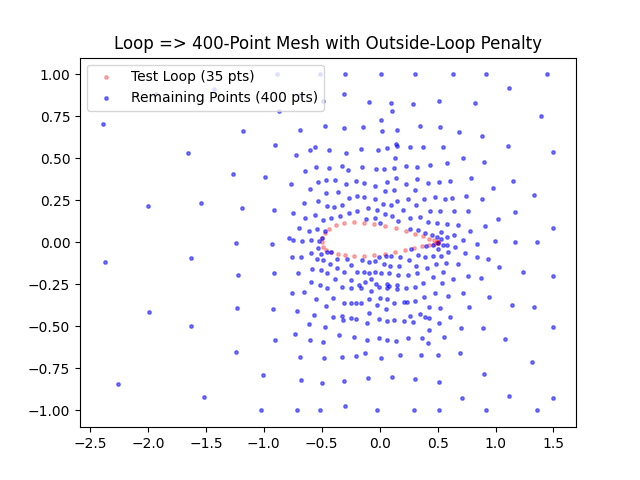}
        \label{fig:2220_non_stand_predict}
    }
    \subfloat[]{
        \includegraphics[width=0.5\linewidth]{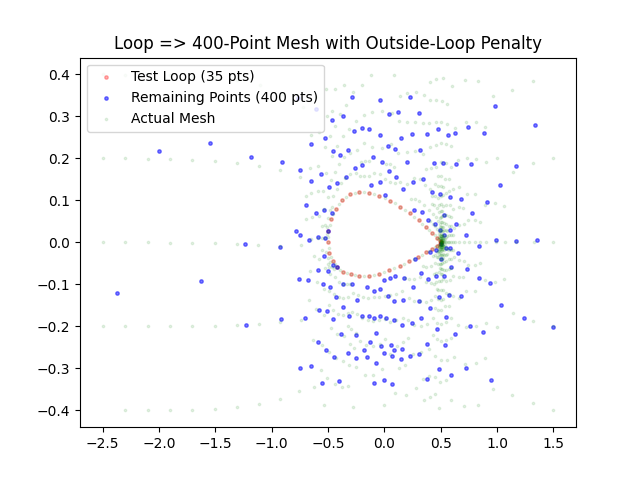}
        \label{fig:2220_non_stand_compare_400}
    }
    \vspace{-2pt}
    \subfloat[]{
        \includegraphics[width=0.5\linewidth]{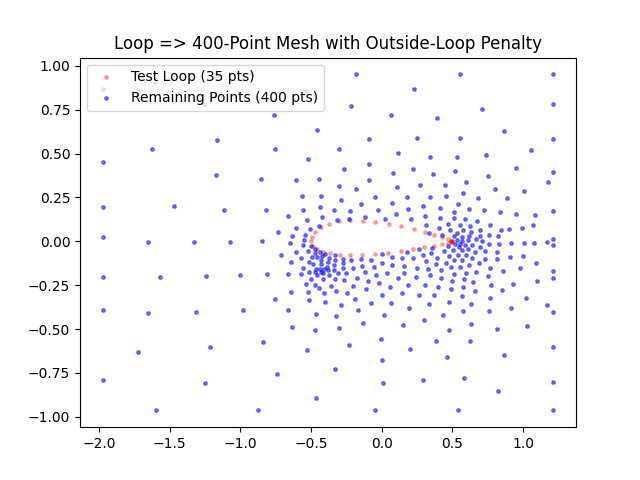}
        \label{fig:2220_stand_ori}
    }
    \subfloat[]{
        \includegraphics[width=0.5\linewidth]{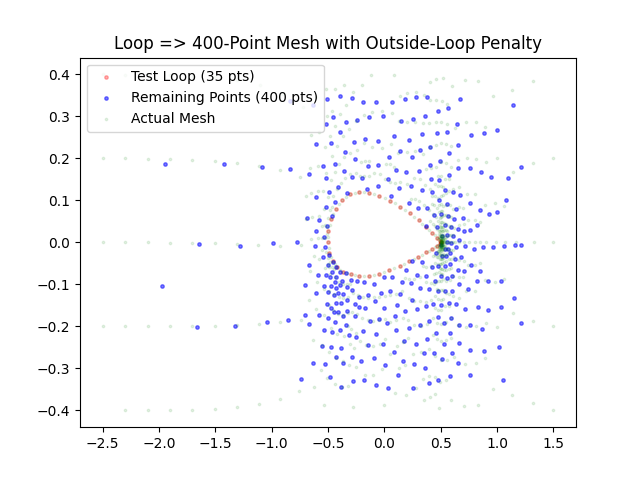}
        \label{fig:2220_stand_compression_compare}
    }
\caption{
Effect of standardisation on mesh prediction.
(a) shows the target CFD mesh; 
(b) displays its standardised counterpart. 
(c) and (d) are predicted meshes generated without standardisation, where (d) is a zoom-in of (c) focused on the core region. 
(e) presents the prediction obtained in the full standardised coordinate space (with $y \in [-2.5, 2.5]$), while (f) shows the result of restricting the prediction range to $y \in [-1, 1]$ during training, effectively compressing the spatial domain. 
Note that when mapped back to the original coordinate system, the output in (f) only occupies a narrow band of $y \in [-0.4, 0.4]$, leading to denser but spatially constrained predictions. 
This figure highlights the trade-off between prediction coverage and training stability introduced by spatial clamping.
}

    \label{fig:standardisation}
    \end{figure}

In this experiment, we first evaluate the effect of standardisation on mesh prediction. Figure \ref{fig:comparison} provides a comparative visualisation. The reference CFD mesh is shown in Figure \ref{fig:2220_ori_target} while the corresponding standardised version is displayed in Figure \ref{fig:2220_stand_target}. Figure \ref{fig:2220_stand_ori} shows the prediction result using a model trained on standardisation inputs and then mapped back to the original coordinates. In contrast, Figure \ref{fig:2220_non_stand_predict} illustrates a direct prediction without any standardisation. \\

We further compare the predictive behaviour under both conditions. Without standardisation, the predicted mesh tends to concentrate around the airfoil, as seen in Figure \ref{fig:2220_non_stand_predict}, but shows limited spatial diversity and stability during training. When standardisation is applied, the input and output coordinates are scaled to zero mean and unit variance, allowing the network to learn more uniformly distributed representations across the domain. This leads to a more structured and smoother prediction, as evident in Figure \ref{fig:2220_stand_ori}. \\

Additionally, we assess how clamping the standardised space affects model output. Figure \ref{fig:2220_stand_compression_compare} shows a prediction generated by restricting the vertical (y-axis) prediction range to $[-1, 1]$ in the standardised space, effectively focusing the network’s attention on the airfoil region. When mapped back to the original scale, this corresponds to a narrow vertical band $[-0.4, 0.4]$, as illustrated in Figure \ref{fig:2220_stand_compression_compare}. To enable fair comparison, Figure \ref{fig:2220_non_stand_compare_400} and Figure \ref{fig:2220_stand_compression_compare} present zoomed-in views of both non-standardised and compressed standardised predictions in the same region. \\


\subsubsection{Train \& Loss}

\begin{figure}[!h]
    \centering
    \includegraphics[width=0.5\linewidth]{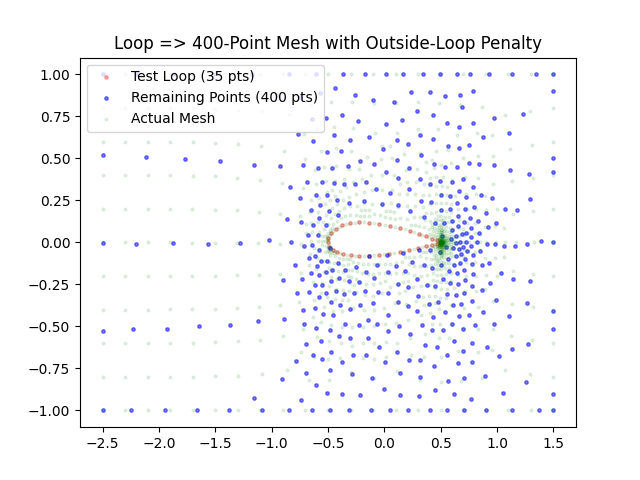}
    \caption{
Comparison between the predicted mesh (blue) in standardisation space and the true CFD mesh (green) with a 400-node target without standardisation. The red points represent the fixed airfoil loop input. The predicted nodes follow the spatial trend of the ground-truth distribution while remaining outside the loop boundary.
}
    \label{fig:comparison}
\end{figure}

To further assess the quality and structure preserving capability of the predicted mesh, we compare the generated node distribution against the true mesh obtained from CFD simulation. Figure \ref{fig:comparison} illustrates a representative case where the network generates 400 mesh nodes based on a fixed airfoil loop input. The red points denote the 35-point loop input, the blue points are the predicted mesh, and the green points represent the ground-truth CFD mesh. \\

\begin{figure}[!h]
    \centering
    \subfloat[1-0]{
        \includegraphics[width=0.5\linewidth]{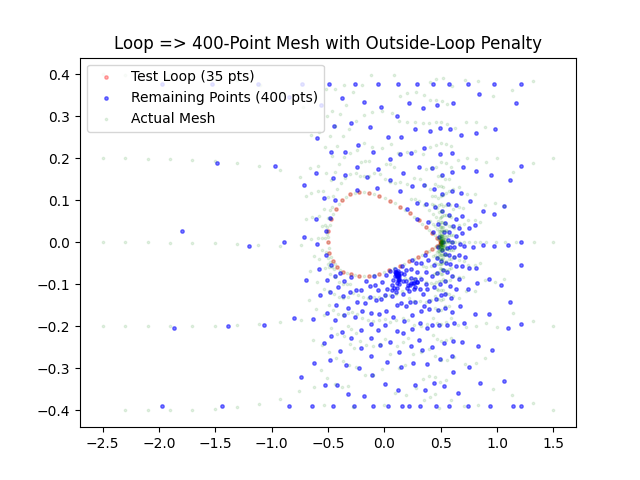}
        \label{fig:400_0_2220}
    }
    \subfloat[1-1]{
        \includegraphics[width=0.5\linewidth]{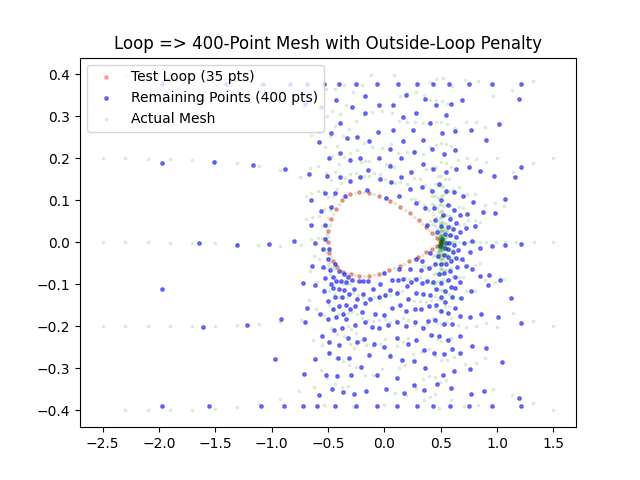}
        \label{fig:400_1_2220}
    }\\[-2pt]
    \subfloat[1-2]{
        \includegraphics[width=0.5\linewidth]{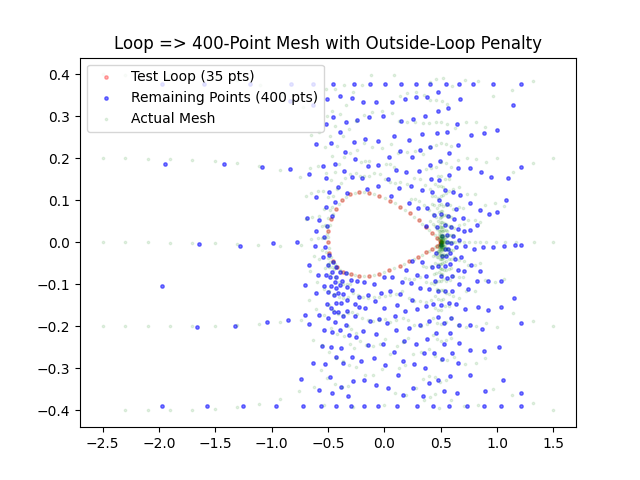}
        \label{fig:400_2_2220}
    }
    \subfloat[1-3]{
        \includegraphics[width=0.5\linewidth]{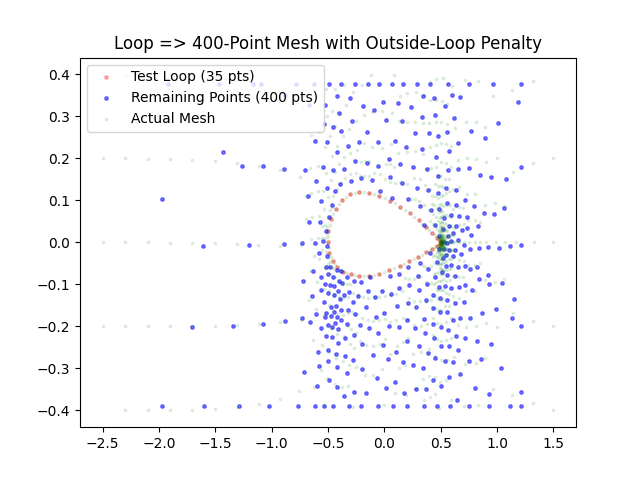}
        \label{fig:400_3_2220}
    }
    \caption{
    Mesh prediction results under varying Chamfer–Repulsion loss ratios. From left to right, the repulsion weight increases from 0 to 3 while keeping Chamfer loss fixed. A balance between structure alignment and spatial dispersion is achieved at intermediate settings.
}
    \label{fig:diff-repulsion}
\end{figure}

The visual comparison in Figure \ref{fig:diff-repulsion} presents the effects of varying the relative weight between Chamfer distance and repulsion loss on mesh generation. From left to right, the repulsion loss weight increases from 0 to 3 while the Chamfer loss remains fixed at 1. As observed, the configuration with no repulsion loss (1-0) leads to high clustering and uneven spacing of mesh nodes. Introducing a moderate repulsion weight (1-1 and 1-2) significantly improves the spatial dispersion of points while preserving adherence to the target shape. Further increasing the repulsion to 1-3 maintains reasonable structural alignment but shows marginal improvements in dispersion. \\


\begin{figure}[!h]
    \centering
    \subfloat[300 nodes]{
        \includegraphics[width=0.5\linewidth]{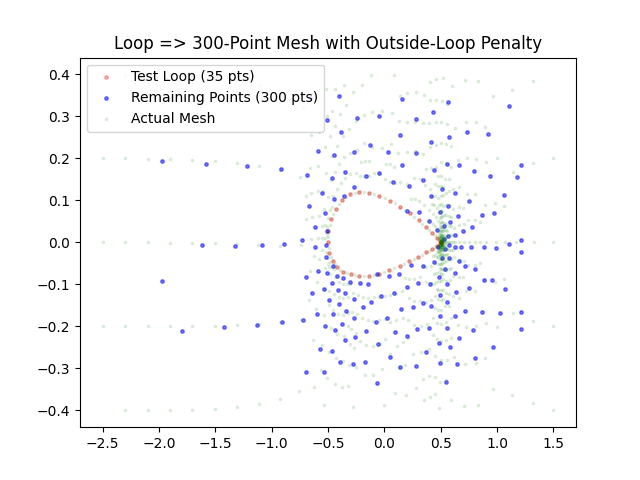}
        \label{fig:300_stand_sample}
    }
    \subfloat[400 nodes]{
        \includegraphics[width=0.5\linewidth]{Sections/media/loop2mesh_2_400_2220_ori_ori.png}
        \label{fig:400_stand_sample}
    }\\[-2pt]
    \subfloat[500 nodes]{
        \includegraphics[width=0.5\linewidth]{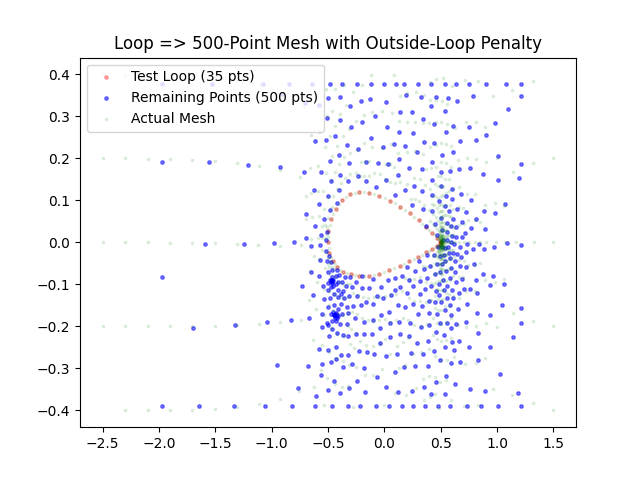}
        \label{fig:500_stand_sample}
    }
    \subfloat[700 nodes]{
        \includegraphics[width=0.5\linewidth]{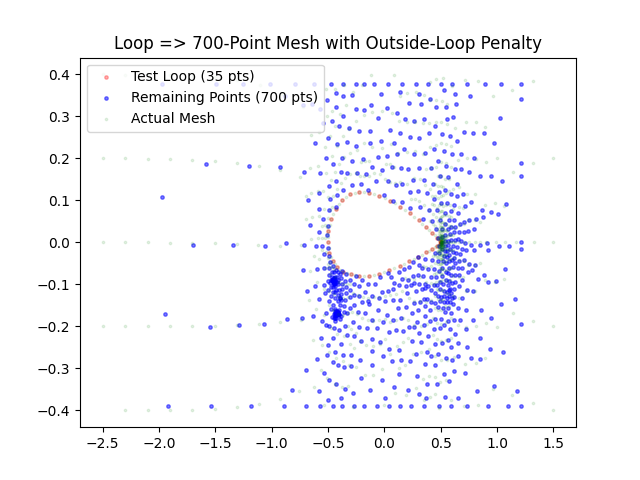}
        \label{fig:700_stand_sample}
    }
    \caption{Mesh prediction results with different target node quantities under a fixed Chamfer–Repulsion loss ratio of 1:2 with clamping the standardised space. The red curve represents the input airfoil boundary (35 points), while the blue dots show the predicted mesh nodes. Subfigures (a) to (d) show the predicted meshes with 300, 400, 500, and 700 nodes, respectively. As the target count increases, the mesh becomes progressively denser near the airfoil and captures more structural details around the wake region.}
    \label{fig:diff-nodes-stand}
\end{figure}

\begin{figure}[!h]
    \centering
    \subfloat[400 nodes]{
        \includegraphics[width=0.5\linewidth]{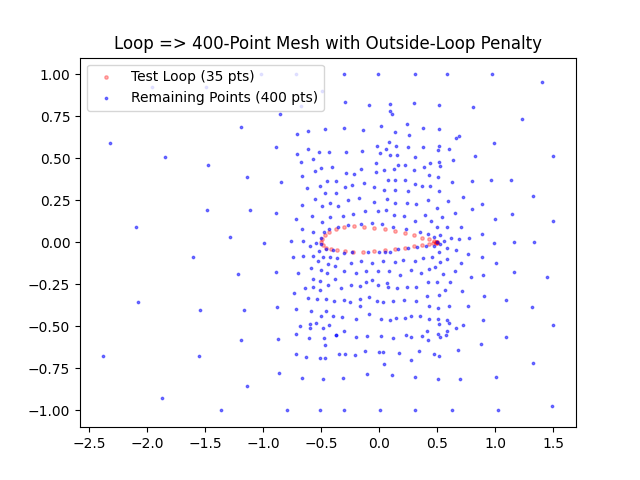}
        \label{fig:400-sample}
    }
    \subfloat[500 nodes]{
        \includegraphics[width=0.5\linewidth]{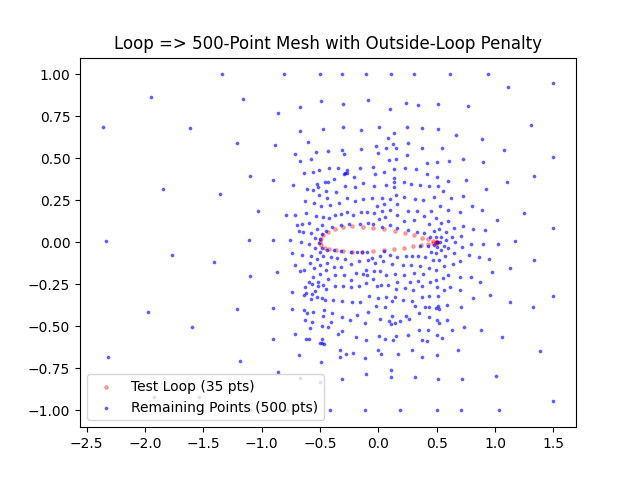}
        \label{fig:500-sample}
    }\\[-2pt]
    \subfloat[700 nodes]{
        \includegraphics[width=0.5\linewidth]{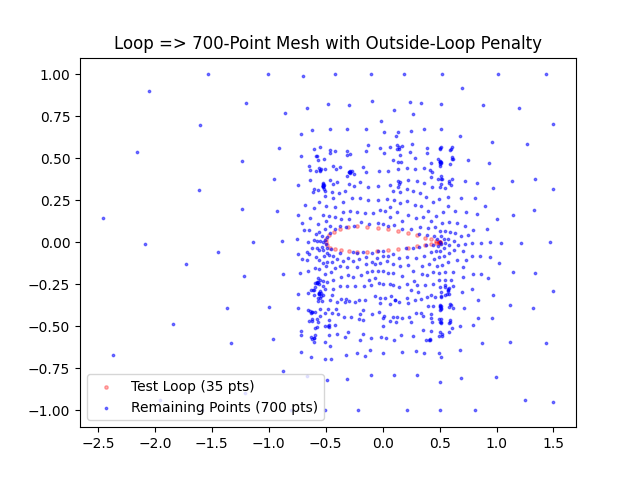}
        \label{fig:700-sample}
    }
    \subfloat[1000 nodes]{
        \includegraphics[width=0.5\linewidth]{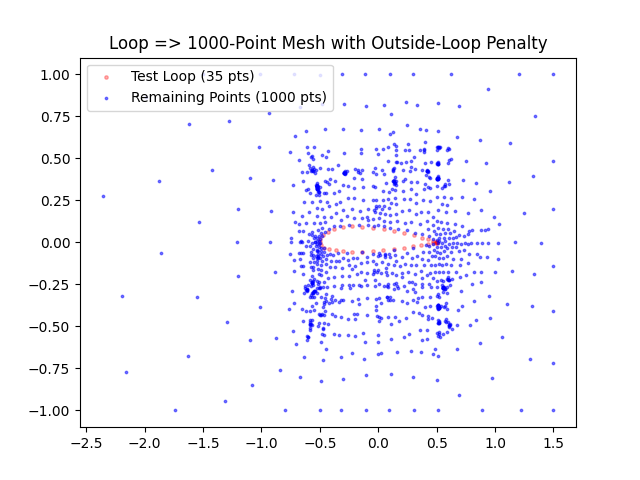}
        \label{fig:1000-sample}
    }
    \caption{
Predicted mesh distributions under different target point counts without standardisation. The red curve denotes the fixed airfoil boundary (35 points), and the blue dots represent the predicted mesh nodes. As the target count increases from 400 to 1000, where the ratio of repulsion loss is 3.5, the mesh becomes progressively denser and expands outward, particularly around the sides and wake of the airfoil.
}
    \label{fig:diff-nodes-no-stand}
\end{figure}

To evaluate the scalability and spatial consistency of the mesh prediction network, we compare the predicted node distributions under varying target point counts, as illustrated in Figures \ref{fig:diff-nodes-no-stand} and \ref{fig:diff-nodes-stand}. In the non-standardised setting, the predicted points tend to cluster unevenly and occasionally extend into regions far from the airfoil, especially as the target count increases. This is evident in the 1000-point case, where a significant portion of the predicted nodes diverges from the expected aerodynamic flow domain. In contrast, when trained with standardised coordinates and proper range constraints, the predicted mesh points maintain a concentrated distribution around the airfoil. The increase in node count from 300 to 700 results in a progressively denser and more detailed mesh, particularly around the leading and trailing edges. Furthermore, These observations demonstrate that standardisation not only enhances prediction stability but also ensures the network generates more structurally coherent meshes as the resolution scales up.

\subsubsection{KL Value Evalutor}
The KL divergence results, shown as Table \ref{tab:nonstand_kl} and Table \ref{tab:stand_kl} provide insight into the distributional fidelity of predicted mesh points under different training configurations. For the standardised model, which compresses the predicted output range, evaluation focuses on the central region of the mesh. In this domain, the standardised model achieves the lowest KL values across all repulsion ratios, with a minimum value of 0.1051 observed at a repulsion-to-alignment loss ratio of 1:1. This indicates strong local fitting and dense alignment in the critical area around the airfoil. \\

In contrast, the non-standardised model, which generates points over the full spatial domain, exhibits consistently lower KL divergence in the global region, particularly at lower loss ratios. The lowest full-domain KL value of 0.1314 is achieved without any repulsion loss, suggesting that the unstandardised model better preserves global distribution characteristics. However, its center-region KL scores remain competitive, revealing that it also maintains reasonable local accuracy. \\

\begin{table}
\centering
\begin{tabular}{clllll}
\multicolumn{6}{c}{Non\_stand\_KL}                                                                  \\
\multicolumn{1}{l}{} & nodes & 300               & 400               & 500               & 700      \\
\multicolumn{6}{l}{c:center, w:whole}                                                               \\
\multirow{2}{*}{0}   & c     & 0.200017          & 0.190961          & 0.307312          & 0.494657 \\
                     & w     & \textbf{0.131444} & 0.190961          & 0.307312          & 0.494657 \\
\multirow{2}{*}{1}   & c     & 0.143908          & 0.181469          & 0.216674          & 0.272808 \\
                     & w     & 0.15222           & 0.152614          & \textbf{0.144484} & 0.16599  \\
\multirow{2}{*}{2}   & c     & \textbf{0.139236} & \textbf{0.151854} & 0.187052          & 0.228359 \\
                     & w     & 0.179489          & 0.177582          & 0.178985          & 0.188684 \\
\multirow{2}{*}{3}   & c     & 0.170121          & 0.17044           & 0.179452          & 0.198721 \\
                     & w     & 0.22257           & 0.215234          & 0.208742          & 0.221009
\end{tabular}
\caption{KL Divergence Comparison (Non-Standardised Prediction)}
\label{tab:nonstand_kl}
\end{table}

\begin{table}
\centering
\begin{tabular}{clllll}
\multicolumn{6}{c}{stand\_KL}                                                                       \\
\multicolumn{1}{l}{} & nodes & 300               & 400               & 500               & 700      \\
\multicolumn{6}{l}{c:center, w:whole}                                                               \\
\multirow{2}{*}{0}   & c     & 0.123233          & 0.219242          & 0.249778          & 0.386358 \\
                     & w     & \textbf{0.219443} & 0.457907          & 0.293313          & 0.412856 \\
\multirow{2}{*}{1}   & c     & 0.148327          & \textbf{0.105184} & \textbf{0.107799} & 0.187207 \\
                     & w     & 0.248078          & 0.312936          & \textbf{0.217824} & 0.282804 \\
\multirow{2}{*}{2}   & c     & 0.142875          & 0.120988          & 0.150362          & 0.165737 \\
                     & w     & 0.33661           & 0.30797           & 0.35877           & 0.409944 \\
\multirow{2}{*}{3}   & c     & 0.140483          & 0.134619          & 0.159641          & 0.168004 \\
                     & w     & 0.327698          & 0.30821           & 0.352755          & 0.398102
\end{tabular}
\caption{KL Divergence Comparison (Standardised Prediction with Compressed Output)}
\label{tab:stand_kl}
\end{table}

\section{Discussion}

The experimental results reveal distinct behavioural patterns between models trained with and without coordinate standardisation. Coordinate standardisation enhances the numerical stability during training and promotes tighter spatial coherence around the airfoil. This behaviour is particularly evident when the number of output nodes remains moderate, where the predicted mesh concentrates accurately near the core region. \\

However, as the target resolution increases, the performance of the standardised model shows limitations in terms of spatial distribution diversity. The predictions tend to accumulate in narrow regions, particularly downstream of the airfoil, suggesting reduced generalisation to broader geometric scopes. In contrast, models trained without standardisation, while potentially less precise near the airfoil, exhibit improved scalability and preserve more natural mesh extension across the domain. \\

Another consideration is that standardisation requires a globally consistent transformation of coordinates, which can introduce mismatches when interpreting model outputs back in the original spatial frame. This additional transformation step may lead to minor distortions, particularly when the predicted region extends toward the domain boundaries. \\

Overall, the results suggest that standardisation is well suited to locally focused prediction tasks or low resolution regimes. On the other hand, the non-standardised approach may be preferable for generating higher resolution meshes or capturing larger scale flow structures. The appropriate strategy should therefore be selected based on the specific application requirements regarding resolution, spatial coverage, and model interpretability.\\

As can be seen from table\ref{tab:nonstand_kl} and table\ref{tab:stand_kl}, the bolded values are the smallest and second smallest values, in the usual sense that the smaller the KL value the better, but it can see that the best vertex layout is shown like in figure\ref{fig:1000-sample}, its not the best shown in table\ref{tab:stand_kl}. This is because it can be seen that the vertices in Figure\ref{fig:1000-sample} are the most consistent with the usual cfd mesh distribution of vertices, which is denser and explosively distributed around the airfoil, and the region of the flow field that needs to be focussed on in front of and behind the airfoil is the most densely populated, it is also one of the closest results to a fine mesh as given in before.

\section{Conclusion}
This study introduces a neural framework for generating dense fluid mesh point distributions based solely on sparse airfoil boundary inputs. Despite the simplicity of using a fully connected architecture, the model demonstrates the capacity to learn meaningful spatial patterns and produce physically plausible meshes. The incorporation of Chamfer Distance and Repulsion Loss enables the network to jointly optimize for geometric accuracy and internal consistency. The findings confirm that, under appropriate training constraints, even minimal network structures can achieve reliable mesh generation performance guided by geometric priors.

\newpage

\printbibliography

\end{document}